\pdfoutput=1

\documentclass[11pt]{article}

\usepackage[final]{coling}

\usepackage{times}
\usepackage{latexsym}
\usepackage{enumitem}
\usepackage{booktabs}

\usepackage[T1]{fontenc}

\usepackage[utf8]{inputenc}

\usepackage{microtype}

\usepackage{inconsolata}

\usepackage{graphicx}

\newcommand{\ie}{\emph{i.e., }}
\newcommand{\eg}{\emph{e.g., }}

\title{Counterfactual Debating with Preset Stances for Hallucination \\Elimination of LLMs}

\author{Yi Fang\textsuperscript{1},
	Moxin Li\textsuperscript{2}, 
	Wenjie Wang\textsuperscript{2}\thanks{Corresponding author.},  
	Hui Lin\textsuperscript{3}, 
	Fuli Feng\textsuperscript{1}\footnotemark[1]
 \\
	\textsuperscript{1}University of Science and Technology of China,
	\textsuperscript{2}National University of Singapore, \\
        \textsuperscript{3}Electronic Science Research Institute of China Electronics \\
	\texttt{peterfang@mail.ustc.edu.cn}, \texttt{limoxin@u.nus.edu}, \\ \texttt{\{wenjiewang96,fulifeng93\}@gmail.com}, \texttt{linhui@whu.edu.cn}
}

\begin{document}
\maketitle
\begin{abstract}


Large Language Models (LLMs) excel in various natural language processing tasks but struggle with hallucination issues. 
Existing solutions have considered utilizing LLMs' inherent reasoning abilities to alleviate hallucination, such as self-correction and diverse sampling methods. 
However, these methods often overtrust LLMs' initial answers due to inherent biases. 
The key to alleviating this issue lies in overriding LLMs' inherent biases for answer inspection. 
To this end, we propose a \textbf{C}ounter\textbf{F}actual \textbf{M}ulti-\textbf{A}gent \textbf{D}ebate (CFMAD) framework. 
CFMAD presets the stances of LLMs to override their inherent biases by compelling LLMs to generate justifications for a predetermined answer's correctness. 
The LLMs with different predetermined stances are engaged with a skeptical critic for counterfactual debate on the rationality of generated justifications. 
Finally, the debate process is evaluated by a third-party judge to determine the final answer. 
Extensive experiments on four datasets of three tasks demonstrate the superiority of CFMAD over existing methods. 

\end{abstract}

\section{Introduction}

Large Language Models, especially closed-source ones such as GPT-4 \citep{gpt4} and Gemini \citep{team2023gemini}, have demonstrated state-of-the-art performance across various natural language processing tasks \cite{bubeck2023sparks, zhao2023survey}. 
However, LLMs still struggle with the hallucination problem, \ie occasionally generating unfaithful content \citep{zhang2023siren,bang2023multitask,zheng2023does}. 
Due to the black-box nature of closed-source LLMs, it is difficult for users to directly intervene in or optimize their internal mechanisms to address the hallucination problems. 
Currently, extensive research is investigating how to use LLMs' inherent reasoning abilities to alleviate hallucinations without model intervention \citep{selfreflection, MAD_diverse}.

\begin{figure}[t]
  \includegraphics[width=\columnwidth]{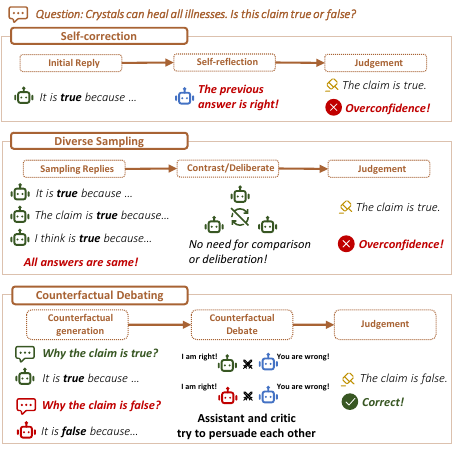}
  \caption{Comparison of CFMAD with self-correction and diverse sampling methods. CFMAD presets stances for LLMs to override their inherent biases.} 
  \label{fig:pipeline}
\end{figure}

Related work of using LLMs' own abilities for hallucination elimination can be categorized into self-correction and diverse sampling methods, which imitate human deep reasoning and broad reasoning to enhance LLMs' reasoning capabilities, respectively \citep{selfcontrast}.
Self-correction methods \citep{selfreflection, madaan2024self} guide LLMs to reflect on and refine their previous answers iteratively. 
Diverse sampling methods \citep{selfcontrast, MAD_fact, wang2023selfconsistency, mielke2022reducing, xiong2023can} first sample multiple initial answers, and then compare or deliberate on the differences among these answers to reach a consistent answer.

While self-correction and diverse sampling methods show potential for improving the output reliability of LLMs, they still have the overconfidence issue \citep{mielke2022reducing, xiong2023can} as illustrated in Figure \ref{fig:pipeline}. 
Self-correction methods may overtrust LLMs' initially generated answers, making it difficult to effectively recognize errors \citep{huang2024large, stechly2023gpt, valmeekam2023can}. 
By contrast, diverse sampling methods may repeatedly generate the same incorrect answers due to LLMs' inherent biases and beliefs \citep{wang2024rethinking}, limiting LLMs to contrast and deliberate on other possible answers. 
We believe that a key reason for the above overconfidence issue is that these methods do not intervene in the LLMs' answer-generation process, allowing LLMs to refine or sample diverse answers according to their own biases and beliefs. 

The main challenge in addressing the overconfidence issue is to override LLMs' inherent biases and beliefs, compelling them to inspect answers they would not normally consider.
To achieve this, we consider presetting different stances for LLMs, allowing LLMs to imagine each answer as correct in each round of reasoning, and then generate the reasons why the answer is valid. 
By overriding the LLM's original beliefs with this new mindset, we can regulate LLMs to assess the possibility of each answer being correct. 
Thereafter, we can eliminate the incorrect answers by reflecting the generated reasons for all answers.

To this end, we propose a \textbf{C}ounter\textbf{F}actual \textbf{M}ulti-\textbf{A}gent \textbf{D}ebate (CFMAD) framework comprising two key stages: abduction generation and counterfactual debate. In the abduction generation stage, LLMs are tasked with producing potential correct reasons for a predetermined answer. Subsequently, in the counterfactual debate stage, a structured debate method is employed to assess these abductions and ascertain the sole correct response. Specifically, we introduce a critic who questions the validity of each generated abduction, and prompt the LLM to defend its position in a debate with the critic. The deliberation is then presented to an impartial third-party judge for final adjudication. Extensive experiments spanning fact-checking, reading comprehension, and commonsense reasoning tasks validate the effectiveness of CFMAD over existing benchmarks across four datasets. We release our code and data at \url{https://github.com/Peter-Fy/CFMAD/}.

The contributions of this work are threefold:
\begin{itemize}[leftmargin=*]
    \item We propose to preset various stances for LLMs, overriding their inherent biases and beliefs to address the overconfidence issue of LLMs. 
    
    \item We propose a CFMAD framework, which instructs LLMs to generate abduction with preset stances and then conduct counterfactual debate to eliminate incorrect answers. 
    
    \item We conduct extensive experiments on three generative tasks with four datasets, validating the effectiveness of CFMAD. 
    
\end{itemize}

\section{Preliminary Experiments}



We formulate methods for self-correction and diverse sampling, and subsequently conduct a quantitative experiment to expose the overconfidence issue prevalent in both approaches. 

\subsection{Problem Definition}

\paragraph{Self-correction.} Self-correction methods involve two steps: reflection and refinement~\citep{selfreflection}. 
Given a question $q$ and $R_0 = LLM(q)$ representing the initial response of an LLM, self-correction methods further instruct the LLM to reflect on the initial response $R_0$ and generate feedback by $F = LLM(q,R_0)$. 
Given $R_0$ and $F$, the LLM then generates a revised answer in the refinement stage, denoted as $R_1 = LLM(q, R_0, F)$. 

\paragraph{Diverse Sampling.} Diverse Sampling methods usually involve three steps: sampling, deliberation, and judging~\citep{selfcontrast}. 
First, $N$ initial responses are sampled by: $R_0^i = LLM(q, \theta_i), i \in [1,N]$. 
Here $\theta_i$ represents settings such as improving temperature or using different prompts, which are widely used in diverse sampling to enhance the diversity of responses. 
In the following deliberation stage, each response is refined by contrasting with other responses, thereby improving the initial responses: $R_1^i = LLM(q, R_0^i, \{R_0^{j \neq i}\})$. 
Finally, a judging process is employed to determine the final answer $R_f = judge(R_1^1, R_1^2, ..., R_1^N)$. 

However, the LLM with self-correction or diverse sampling face the issue of overconfidence. Formally, the LLM with self-correction tends to overtrust the initial response $R_0$, resulting in $R_1$ having the same error as $R_0$ \citep{selfcontrast}. Meanwhile, for diverse sampling, the incorrect answer might repeat in $\{R_0^1, R_0^2, ..., R_0^N\}$, resulting in the deliberation and judging stages potentially accepting such an incorrect answer.





\subsection{Investigation of Overconfidence} \label{section22}

To investigate the overconfidence issue, we conduct some preliminary experiments on the representative self-correction and diverse sampling methods. 

\paragraph{Testing Methods.} 
We evaluate four representative methods and count the number of testing samples exhibiting the overconfidence issue as follows:
\begin{itemize}[leftmargin=*]
\item{
\textbf{Self-reflection} \citep{selfreflection}: This method instructs the LLMs to reflect on an initial answer and subsequently provide feedback, asking the LLM to refine and generate a revised response based on this feedback. 
If the revised answer for a testing sample remains the same as the initial incorrect response, we treat it as a sample with the overconfidence issue. 
}

\item{
\textbf{Self-consistency} \citep{wang2023selfconsistency}: This approach samples multiple initial answers using the same prompts, followed by a voting process to determine the final answer. 
We implement it by sampling seven initial answers and consider a test sample as an overconfidence sample if six out of the seven responses are identically incorrect. 
}

\item{
\textbf{Self-contrast} \citep{selfcontrast}: In this method, three initial answers are generated by the LLMs using self-generated, varying prompts. 
These answers are then contrasted to derive the final answer. 
If all three initial responses are the same incorrect answers for a given testing sample, it is regarded as an overconfidence sample.
}
\item{
\textbf{MAD} \citep{MAD_fact}: This strategy involves sampling multiple initial answers from different agents and using a debate process to decide on the final answer. 
Similarly, an overconfidence sample is defined as that three initial responses are the same and incorrect. 
}
\end{itemize}

\begin{figure}[t]
  \includegraphics[width=\columnwidth]{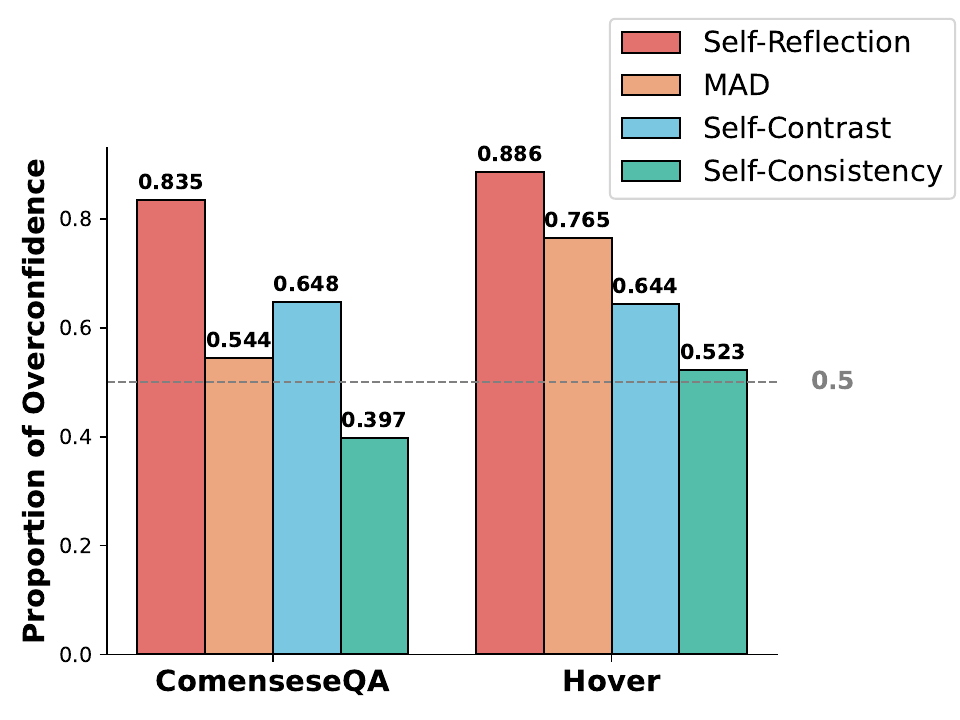}
  \caption{Proportion of the overconfident answers among all incorrect answers.} 
  \label{fig:preliminary_result}
\end{figure}

\paragraph{Results.} 
We assess the overconfidence issue by applying these methods to a representative LLM, GPT-3.5-turbo~\citep{NEURIPS2022_b1efde53}, on the CommonsenseQA \citep{talmor2019commonsenseqa} and Hover \citep{hover} datasets. 
We calculate the proportion of incorrect answers attributable to overconfidence among all incorrect cases. 
As shown in Figure~\ref{fig:preliminary_result}, for self-reflection, MAD, and Self-Contrast, more than half of the errors are caused by overconfidence. 
For Self-consistency, although the overconfidence issue is alleviated due to the increase in sample number and temperature, approximately 40\% of the errors are still caused by the overconfidence of LLMs. This validates the severity of the overconfidence issue in existing self-correction and diverse sampling methods. 

A key reason for the overconfidence issue of LLMs might be that self-correction and diverse sampling methods do not intervene in the LLM's answer-generation process, permitting LLMs to refine and sample diverse answers based on their inherent biases and beliefs. 
Consequently, LLMs tend to trust the initial incorrect answer, hindering the consideration of alternative potential answers. 



\section{Counterfactual Multi-agent Debate}

\begin{figure*}[!ht]
    \centering
    \includegraphics{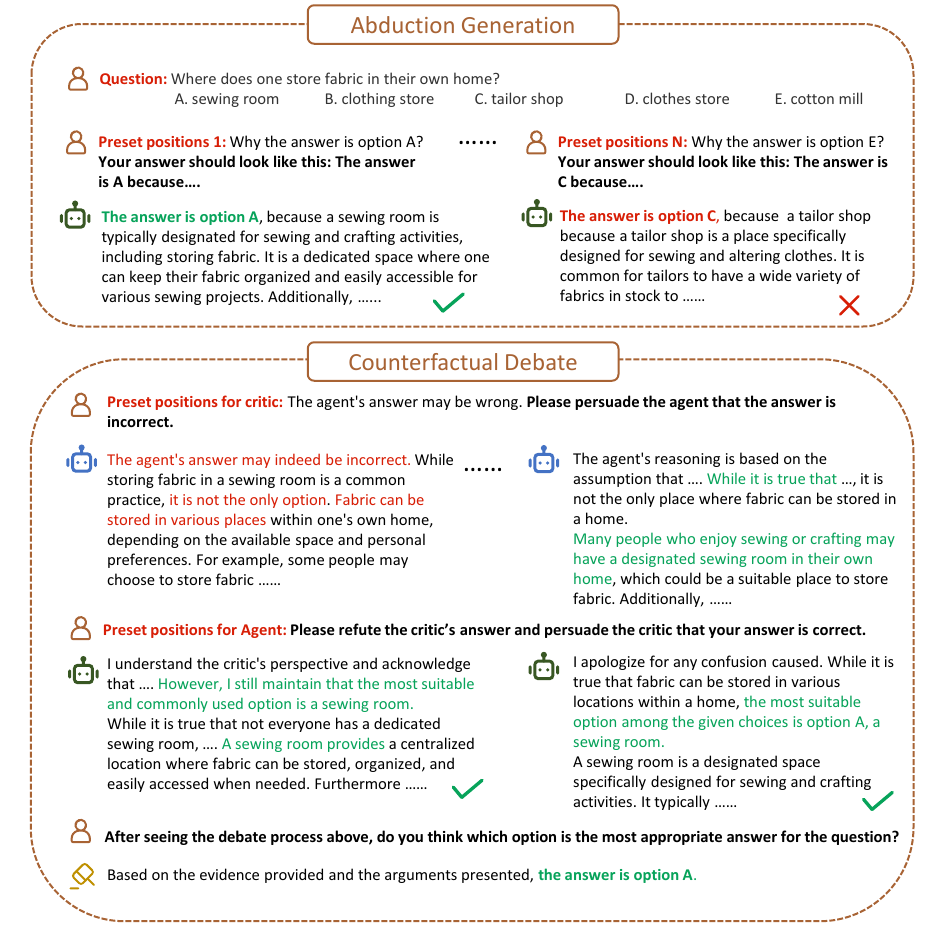}
    \caption{Illustration of CFMAD framework with two stages. In the abduction generation stage, we initialize multiple LLM agents, each configured to assume a specific answer is correct and to generate supporting abductions. In the subsequent counterfactual debate stage, each agent is challenged by a critical evaluator for debating. The debating processes are assessed by a third-party judge for final adjudication.} 
    \label{fig:enter-label}
\end{figure*}

To address the overconfidence issue, the key lies in overriding the inherent biases and beliefs of LLMs for answer generation. 
To achieve this, 
we consider initially configuring the LLMs with various stances, allowing them to hypothesize the correctness of each possible answer and uncover the underlying rationale of each answer. This approach compels LLMs to inspect all potential answers, reducing their reliance on inherent biases and beliefs. Subsequently, we can critically assess the potential rationales to identify the correct answer.

To this end, we propose a CFMAD framework comprising two sequential stages: abduction generation and counterfactual debate, depicted in Figure~\ref{fig:enter-label}. 
In the abduction generation phase, we initialize multiple LLM agents and configure each one to adopt a predetermined stance, assuming a specified answer is correct. Subsequently, these agents are instructed to generate abductions, \ie potential correct reasons for the given answer. 
In the counterfactual debate phase, we create an adversarial debate scenario. Each abducting agent, adopting a predetermined answer as correct, faces a critical evaluator tasked with challenging the validity of the abductions generated by the agent. Meanwhile, the abducting agent is directed to defend its position on the abduction correctness. 
Eventually, the deliberations between each agent-critic pair are presented to a third-party judge to deliver the final adjudication. 





\subsection{Abduction Generation}

Prior studies have illustrated that LLMs exhibit proficiency in counterfactual reasoning \cite{nguyen2024llms, bhattacharjee2024zero}, thereby allowing them to engage in reasoning with predetermined stances. 
Specifically, given a possible answer $a_i$, we preset the LLMs' stance with $a_i$ by the following prompt: 
\vspace{-12pt}
\begin{center}
\fcolorbox{black}{gray!6}{\parbox{0.98\linewidth}{
Why is $a_i$ the correct answer? Your answer should look like this: The answer is $a_i$ because ...}} 
\end{center}
Even if $a_i$ is incorrect, the LLM can still follow our instructions to perform counterfactual reasoning to generate plausible justifications.

Drawing from this insight, CFMAD assigns the LLM agents the task of generating abductions for each potential answer. Concretely, as depicted in Figure~\ref{fig:enter-label}, when presented with a set of possible answers $\{a_1,a_2,...,a_M\}$, we activate multiple abducting agents. Each agent is tasked with assuming that a specific answer $a_i$ is correct and then generating the corresponding abduction $r_i$. 

\subsection{Counterfactual Debate}
Among the generated abductions $\{r_1,r_2,...,r_M\}$, only one is factual, while the remainder are incorrect justifications. Hence, we introduce a counterfactual debate mechanism to discern the correct answer from the pool of abductions. 
Specifically, for each abducting agent $g_i$, who is preset with the position that $a_i$ is correct, we introduce a critic evaluator to challenge the correctness of $a_i$. 
By showing the agent's abduction $r_i$ to critic $c_i$ and instructing the critic with a prompt like: 
\vspace{-12pt}
\begin{center}
\fcolorbox{black}{gray!6}{\parbox{0.98\linewidth}{
The agent's answer may be wrong. Please persuade the agent that the answer is incorrect.}} 
\end{center}
Simultaneously, we preset the stance of $g_i$, ensuring it firmly believes in the correctness of its answer and addresses challenges from the critic. For instance, we provide $g_i$ with a prompt such as:
\vspace{-12pt}
\begin{center}
\fcolorbox{black}{gray!6}{\parbox{0.98\linewidth}{
Please refute the critic’s answer and persuade the critic that your answer is correct.}}
\end{center}
With the aforementioned configuration, we orchestrate an adversarial debate scenario for each agent-critic pair.

The abduction $r_i$ for an incorrect answer $a_i$ inevitably incorporates numerous fabricated reasoning processes and factually incorrect elements. The adversarial debate process will help to unveil the errors or unreasonable justifications in $r_i$. 

After multi-round debating, we present the debate process of all agent-critic pairs to a third-party judge, implemented as another LLM. This enables the judge to meticulously analyze and juxtapose the varied debate trajectories, thereby discerning the final answer. An example prompt for the third-party judge can be found in the Appendix~\ref{prompt2}.




\section{Experiments}
In this section, we conduct extensive experiments on the widely studied fact-checking, reading comprehension, and commonsense reasoning tasks.
\begin{table*}[t]
\centering
\begin{tabular}{lcccccc}
\hline
\textbf{Method} & \textbf{Hover 3-hop} & \textbf{Hover 4-hop} & \textbf{BoolQ} & \textbf{CosmosQA} & \textbf{CommenseQA} \\
\hline
CoT & 0.6108 & 0.5886 & 0.7767 & 0.7833 & 0.7467 \\
Self-Reflection & 0.5986 & 0.5813 & 0.7728 & 0.7867 & 0.7567 \\
Self-Consistency & 0.6342 & 0.6044 & 0.8033 & 0.8067 & \underline{0.7733} \\
MAD & \underline{0.6476} & 0.6069 & 0.8020 & 0.7933 & 0.7700 \\
Self-Contrast & 0.6359 & \underline{0.6178} & \underline{0.8267} & \underline{0.8133} & 0.7633  \\
\hline
CFMAD (Ours) & \textbf{0.6757} & \textbf{0.6361} & \textbf{0.8366} & \textbf{0.8267} & \textbf{0.7933} \\
\hline
\end{tabular}
\vspace{-0.2cm}
\caption{Overall performance comparison on all experiment datasets. Bold font and underline indicate the best and second-best performance, respectively.}
\label{tab:main_resut}
\vspace{-0.2cm}
\end{table*}

\subsection{Experimental Setup}

%
\paragraph{Datasets.} We conduct experiments on four datasets: Hover \citep{hover}, BoolQ \citep{clark2019boolq}, CosmosQA \citep{huang2019cosmos}, and CommonsenseQA \citep{talmor2019commonsenseqa}. Hover and BoolQ are binary prediction tasks with only true or false answers. CosmosQA and CommenseQA are multi-choice tasks with 4 and 5 options, respectively. 
Note that we split Hover into two subsets named Hover 3-hop and Hover 4-hop with questions requiring 3 and 4 steps of reasoning, respectively. Comparison between Hover 3-hop and Hover 4-hop might reveal the influence of problem difficulty on method effectiveness since more complex questions typically require more reasoning hops. 
More details about these datasets can be found in the Appendix \ref{datesetDetails}.

%
\paragraph{Baselines.} As introduced in Sections~\ref{section22}, we compare CFMAD with the four baselines: Chain-of-thought (CoT) prompting \citep{wei2022chain}, Self-Reflection \citep{selfreflection}, Self-Consistency \citep{wang2023selfconsistency}, Self-Contrast \citep{selfcontrast}, MAD \citep{MAD_fact}.

%
\paragraph{Implementation Details.} 
The implementation detail involves three key factors: backbone, prompt, and inference temperature. For all compared methods, we use GPT-3.5-turbo-0613\footnote{\url{https://chatgpt.com/}.} as our backbone LLM and present their prompts in Appendix~\ref{prompt1} and \ref{prompt2}. As to the inference temperature, we set it to 0.2 in most methods for the sake of fair comparison. The only exception is Self-Consistency, where we follow the original paper and set the temperature to 1 since the method requires high diversity of samples~\citep{wang2023selfconsistency}.

%
\paragraph{Evaluation Metrics.} For binary prediction datasets, \ie Hover and BoolQ, we follow the previous work \citep{wang2023explainable} and adopt the macro-F1 score as the evaluation metric. As to multi-choice datasets, \ie CosmosQA and CommenseQA, we report accuracy following previous work \citep{wang2023gemini}.


%
\subsection{Performance Comparison}
Table~\ref{tab:main_resut} shows the performance of the compared methods on all datasets. From the table, we have the following observations:
\begin{itemize}[leftmargin=*]
    \item In all cases, CFMAD outperforms all baselines, showing stronger reasoning capabilities. Such performance gain indicates the effectiveness of the abduction generation and counterfactual debate mechanism.
    \item Among all self-correction and diverse sampling methods, Self-Reflection performs the worst in all cases, sometimes even worse than CoT. Given that Self-Reflection encounters the most severe overconfidence issue (as shown in Figure~\ref{fig:preliminary_result}), we postulate that such inferior performance is due to overconfidence.
    \item While Self-Consistency exhibits lower levels of overconfidence than Self-Contrast by providing answers within a wider scope, it does not consistently outperform Self-Contrast across all tasks. This suggests that incorporating diverse perspectives alone does not guarantee superior reasoning outcomes; the effective utilization of these varied viewpoints is crucial for optimal performance.
\end{itemize}

\begin{figure}[t]
\setlength{\abovecaptionskip}{0.1cm}
\setlength{\belowcaptionskip}{-0.30cm}
\includegraphics[width=\columnwidth]{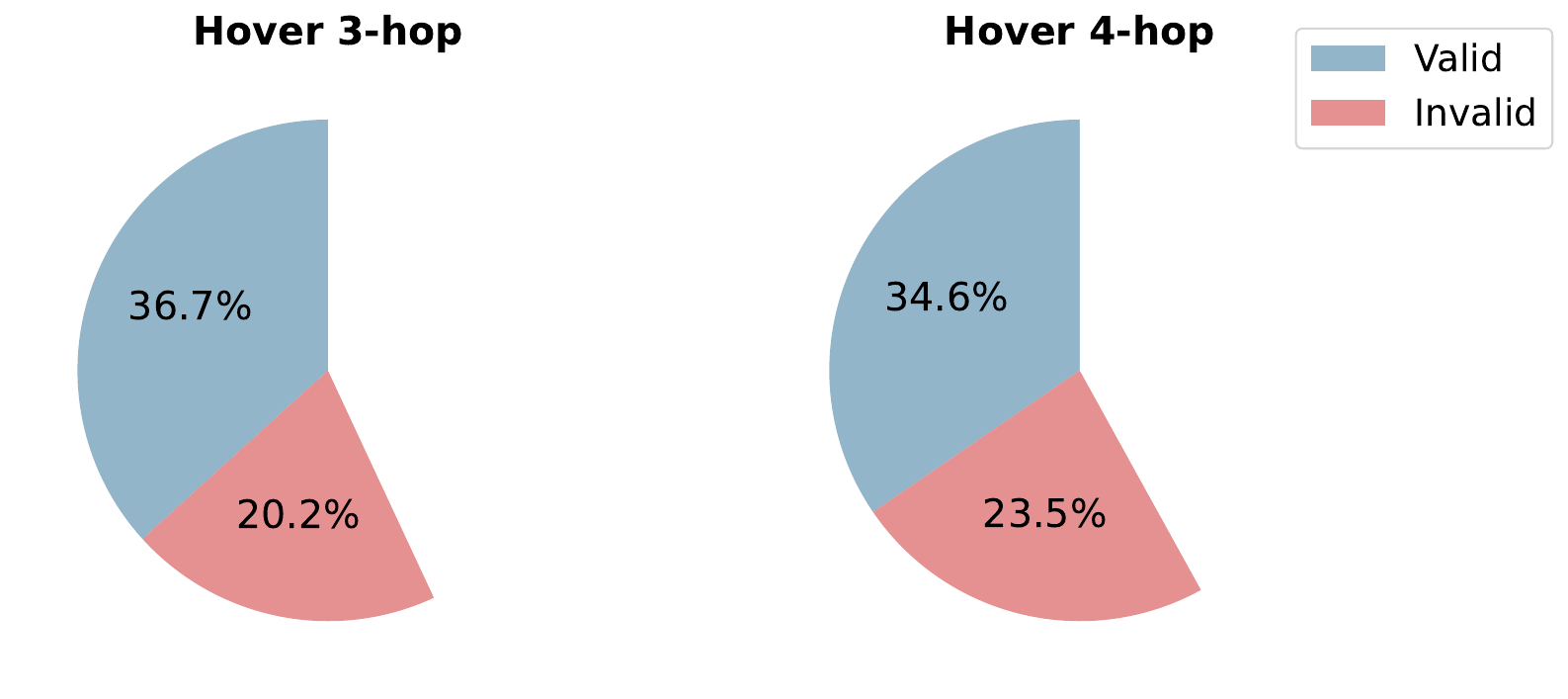}
  \caption{\textbf{Proportion of changes in initial stances.} ``Valid'' means the stances changed from incorrect to correct. ``Invalid'' represents the stances changed from correct to incorrect.}
  \label{fig:change}
\end{figure}

\subsection{In-depth Analysis}
We proceed to analyze the performance enhancement of CFMAD. We posit that the efficacy of CFMAD stems from two key factors: 1) Agents instructed to generate abductions for incorrect answers are more likely to waver and change their stance during the debate process due to the contradictions with factual information. 2) Engaging in counterfactual debates aids judges in distinctly discerning between accurate and inaccurate answers. Subsequently, we undertake experimental investigations to delve into these aspects.

\subsubsection{Counterfactual Answers are More Prone to Change}
As to stance change, we first analyze whether the agents would change their stance even when instructed to maintain their original position. For simplicity, we conduct our analysis using the Hover dataset with binary answers. Specifically, we first ask two agents to generate abductions for both ``True'' and ``False'' answers, respectively. Given that there are only two possible answers, one of these abductions is necessarily factual while the other is counterfactual. 
Given these abductions, we conduct a single round of counterfactual debate. For both agents with factual and counterfactual abductions, we instruct the critic to persuade the agent that their claim is actually incorrect. After that, we present the critic's argument to the corresponding agents and instruct them to maintain their original stance by pointing out the errors in the critic's answer and reiterating your point.
Finally, we observe whether these factual and counterfactual agents would change their stance. 

\begin{figure}[t]
\setlength{\abovecaptionskip}{0.1cm}
\setlength{\belowcaptionskip}{-0.30cm}
\includegraphics[width=\columnwidth]{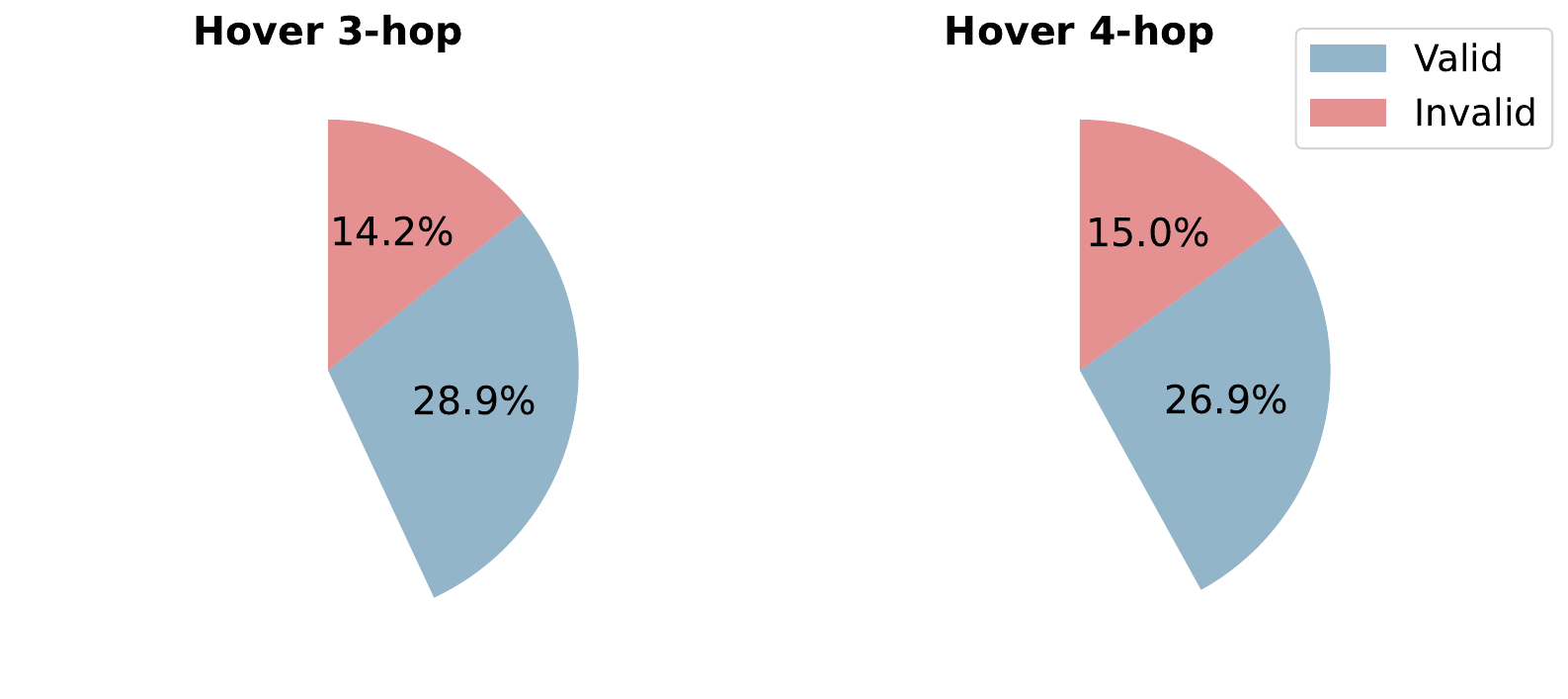}
  \caption{\textbf{The final judgment on inconsistent stances.} ``Valid'' means that the judge makes a correct judgment while ``Invalid'' denotes making an incorrect judgment.}
  \label{fig:judge}
\end{figure}

The results on Hover 3-hop and Hover 4-hop are shown in Figure~\ref{fig:change}. From Figure~\ref{fig:change}, we find that over 50\% of factual and counterfactual agents reached a consensus after one round of counterfactual debate. It means that a significant number of agents were persuaded by the critic, while we instruct these agents to maintain their original stance. Specifically, more than 34\% of the stance changes came from counterfactual agents, which is 10\% higher than the changes from factual agents. We believe this is because counterfactual answers inherently contradict the facts, making it easier for the critic to point out issues and for the agents to realize the problems and subsequently change their stance.

\subsubsection{Counterfactual Debates Contain Additional Clues}

We first analyze the contribution of the counterfactual debate by continuing the previous experiment. For those agents that do not reach a consensus, we present the entire debate process between the critic and the factual and counterfactual agents to a third-party judge. 
The judge then makes a final decision on which stance is more factual. 
As shown in Figure~\ref{fig:judge}, the number of correct judgments was twice that of incorrect judgments, indicating that even if a consensus is not ultimately reached, leveraging the judge to evaluate the counterfactual debate process can still significantly improve the accuracy of the final decision. 

To further investigate the effectiveness of the counterfactual debate, we evaluate several variants of CFMAD, including: 
\begin{itemize}[leftmargin=*]
\item{
\textbf{Direct Judge:} Removing the counterfactual debate and directly presenting the generated abductions to the judge for final decision. 
}
\item{
\textbf{Replace with Self-Reflection:} Replacing the counterfactual debate with self-reflection, where the LLM reflects on each generated abduction. Both the original answer and the reflection process were shown to the judge for final decision. 
}
\item{
\textbf{Replace with MAD:} Replacing the counterfactual debate with three rounds of MAD \citep{MAD_fact}, then presenting the MAD debate process to the judge for the final decision.
}
\end{itemize}
We conduct the ablation experiments on three datasets. For CosmosQA and CommenseQA, we used the same 300 data as in Table~\ref{tab:main_resut}. For Hover 3-hop, we randomly sampled 300 data points due to cost limitations.
The results are shown in Table~\ref{tab:ablation_study}. We can see that our proposed counterfactual debate component outperforms the other control group across all tasks. This demonstrates that the counterfactual debate component helps the judge more effectively determine the correct final answer.

\begin{table}[t]
    \centering
    \setlength{\tabcolsep}{0.8mm}{
    \resizebox{\columnwidth}{!}{
    \begin{tabular}{lccc}
        \hline
        Method & Hover 3-hop & CosmosQA & CommenseQA \\
        \hline
        CFMAD & \textbf{0.6815} & \textbf{0.8267} & \textbf{0.7933} \\
        Direct Judge & 0.6027 & 0.7633 & 0.7500 \\
        Repl. w/ SR & 0.6063 & 0.7800 & 0.7600 \\
        Repl. w/ MAD & 0.6224 & 0.6767 & 0.7200 \\
        \hline
    \end{tabular}
    }
    }
    \caption{Ablation studies on the effectiveness of our counterfactual debate component.}
    \label{tab:ablation_study}
\end{table}

\subsection{Impact of Hyperparameters}
We then investigate the influence of hyperparameters on the effectiveness of CFMAD, including the number of initial counterfactual answers and debate rounds. 
\begin{figure}[t]
\setlength{\abovecaptionskip}{0.1cm}
\setlength{\belowcaptionskip}{-0.30cm}
\includegraphics[width=\columnwidth]{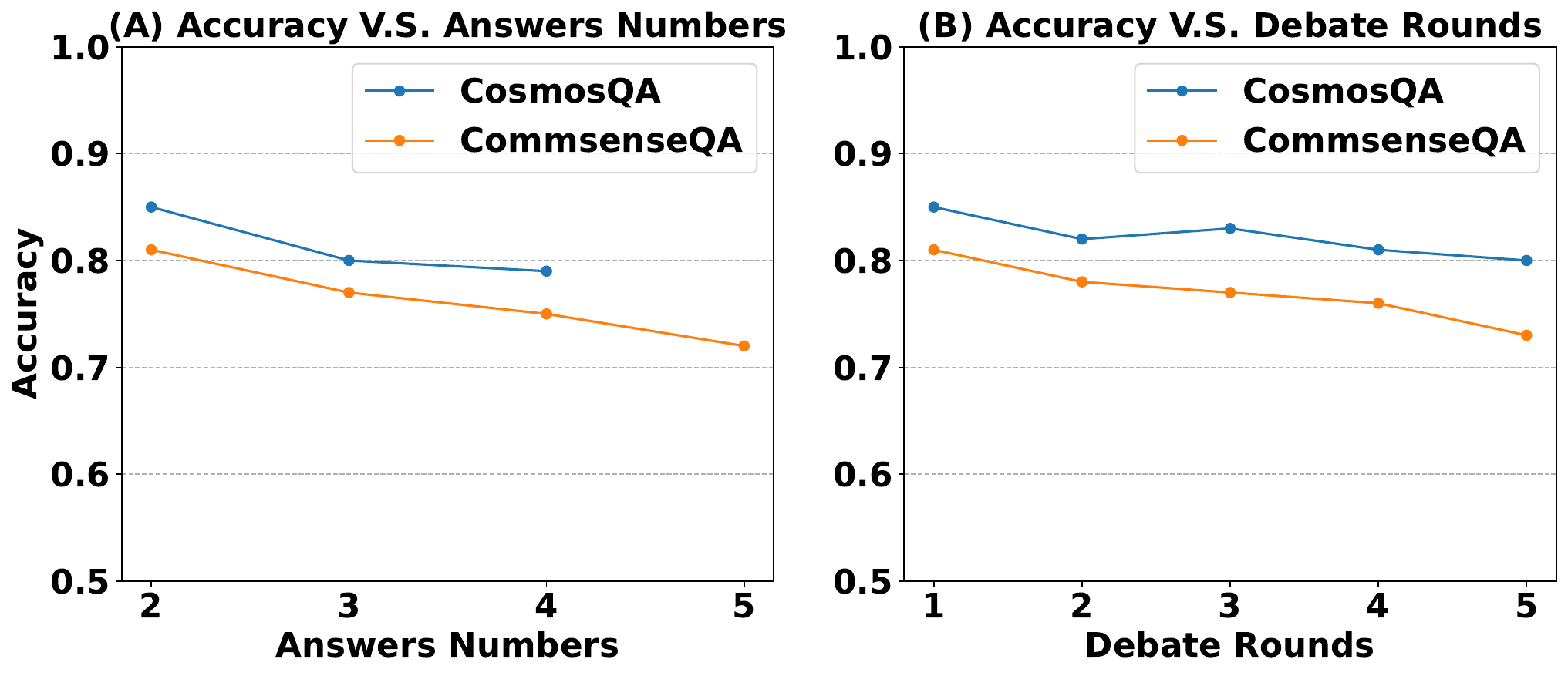}
  \caption{Comparison of different numbers of (A) initial counterfactual answers and (B) debate rounds. }
  \label{fig:onemore}
\end{figure}

%
\paragraph{Number of Initial Counterfactual Answers.} Considering that datasets like CosmosQA and CommenseQA have multiple potential answers, we explore the influence of initial counterfactual answers by increasing the number of sampled stances. Note that directly sampling a few stances from many options may fail to include the correct answer when the initial number of stances is small (\eg 2 out of 5 choices). We thus need to conduct the comparison under the condition that the correct answer is included.
To this end, we use a CoT prompt to generate three answers and select the most frequently occurring answer as the most potential stance. We then randomly sample the remaining stances to complete the initial settings. 
Considering the expensive time and monetary costs, we randomly sampled 100 data from each dataset.
The final result is shown in Figure~\ref{fig:onemore}(A), where the accuracy of the final judgment decreases as the number of initial counterfactual responses increases. We believe this is because the presence of too many incorrect stances can confuse the LLMs. Notably, only two initial counterfactual answers are needed to achieve good results, which also saves time and cost.

\paragraph{Number of Debate Rounds.}
We also test the impact of conducting multiple rounds of counterfactual debate. 
As shown in Figure~\ref{fig:onemore}(B), the accuracy decreases with the increase of debate rounds. 
We speculate that through multiple rounds of debate, LLM-based agents and critics may veer away from our predetermined stances to adhere to the biases in the LLM itself, thereby influencing the efficacy of the debate. As such, we conduct only one-round debate between the agent and critic by default.

\section{Related Work}


\paragraph{Prompting LLM for Better Reasoning.} 
Researchers have made significant progress in improving the reasoning abilities of LLMs through designing better prompting methods. 
These methods often enhance the LLM's reasoning capabilities in either reasoning depth or breadth.
CoT prompting \citep{wei2023chainofthought} guides the model to generate intermediate reasoning steps before arriving at a final answer, thus improving the reasoning depth. 
Self-correction methods \citep{madaan2024self, selfreflection, refiner, xi2024selfpolish} are also typical examples of enhancing LLM reasoning depth. 
They leverage the LLM's self-correction ability, generating feedback by LLM itself to iteratively refine its answers, thereby enhancing its accuracy and reliability. 
Breadth reasoning approaches, on the other hand, involve sampling diverse responses with temperature larger than 0 \citep{wang2023selfconsistency, yoran2023answering} or guiding the LLM to generate responses from different perspectives \citep{huang2024enhancing, selfcontrast}, gathering more diverse insights for the answer. 
This helps to derive the correct answer from the collection of a wider range of potential responses to improve the overall reliability. 

\paragraph{Multi-agent Debate.} 

Recent research has explored how to engage multiple agents of the same model or different models in debates to jointly improve decision-making and reasoning processes \citep{MAD_fact, MAD_diverse, wang2024rethinking}, which can be divided into two modes: collaborative and adversarial. 
In the collaborative mode, each agent provides its own answer to the same question and then refines its answer with reference to the responses of other agents \citep{MAD_fact}. 
This mode may encounter overconfidence issues that the initial responses of most agents arrive at the same incorrect answer. 
In the adversarial mode, for a given answer, two agents are initialized: one believing the answer is correct, and the other believing the answer is incorrect, and they are instructed to debate and challenge each other's response to reach a more precise conclusion \citep{MAD_diverse, wang2024rethinking}. 
The difference between our counterfactual debate and the adversarial debate lies in that they first have the LLM generate a single answer and then conduct a debate about that answer, while we first have the LLM explore multiple answers as thoroughly as possible, and then conduct debates for each of these answers. 
Additionally, existing work also leverages multiple side rationales in LLM reasoning \cite{jung2022maieutic, liu2023score, balepur2023s} which is similar to our abduction, yet not all of them has shown promising results. We incorporate them in the counterfactual debate process and achieve enhanced reasoning. 

\paragraph{Confidence Calibration.}  

Recently, confidence calibration for LLMs has gained significant attention \citep{lin2022teaching, Kuhn_Gal_2023, Huang_Song_2023, tian2023just}. 
The goal of confidence calibration is to obtain LLM's confidence score on its own answer which aligns with the actual answer accuracy. 
However, some studies found that LLMs sometimes 
generate confidence scores that are poorly calibrated and often assign high confidence scores to incorrect answers \citep{shrivastava2023llamas, yang2024confidence, xiong2024llms}. 
Some methods attempt to calibrate the confidence for LLMs through estimating response consistency across multiple perspectives \citep{zhang2024calibrating, wang2024multi}, and various prompting strategies for LLM to self-estimate the confidence \cite{tian2023just, kadavath2022language, li2024think}, where some work also leverages explanation and rationales \cite{li2024think, feng2024don}. 
However, these works mainly aim at improving calibration errors or identifying incorrect answers instead of directly improving the answer accuracy. 

\section{Conclusion}

In this paper, we addressed the overconfidence issue presented in existing self-correction and diverse sampling methods for hallucination elimination in LLM reasoning.
We revealed the overconfidence issues of these two methods through experiments, and pointed out that the overconfidence issue mainly stems from the LLM's inherent biases towards overly favoring a particular answer while lacking sufficient exploration of other potential answers. 
To address this, we proposed the CFMAD framework, which first presets the stance for the LLM, encouraging it to explore as many answers as possible, and then uses counterfactual debate to expose and correct the errors in the incorrect answers.
Empirical results validate the superiority of CFMAD over baselines in mitigating hallucinations. 
In this work, we mainly test CFMAD on binary and multiple-choice questions. In the future, we intend to extend CFMAD to more scenarios with open-ended questions.  


\section*{Limitations}

Our work has the following limitations: First, we require the LLM to generate reasons for each possible answer and conduct debates for each answer, which results in additional computational overhead. 
Secondly, since it is necessary to preset the stance for the LLM, we must identify potential answers. We address this by initially using CoT prompts sampling to generate three possible answers. However, it is worth exploring superior methods to improve the recall rate of correct answers. 

\section*{Ethics Statement}

Our ethical concerns include the following points. First, although we can mitigate LLM hallucinations using CFMAD, the LLM may still produce some inaccurate answers, which could potentially cause harm. Secondly, our experiments are conducted exclusively on English datasets, meaning the applicability of our findings to other languages has not been comprehensively evaluated.

\section*{Acknowledgments}
We thank the reviewers for their constructive feedback.

\bibliography{custom}

\appendix

\section{Complementary Experiments}
\subsection{Token Efficiency Analysis}

We compared the token costs of our method to several baselines on the CommonsenseQA dataset, as shown in Table~\ref{tab:token_costs}. Specifically, We computed the number of prompts used per data sample, the average number of tokens per prompt for a single sample, and the total average token usage per sample. The results indicate that our method's token costs are similar to those of other multi-agent methods, such as MAD and Self-Contrast. Similar results are observed on other datasets.

\begin{table}[ht]
    \centering
    \setlength{\tabcolsep}{0.8mm}{
    \resizebox{\columnwidth}{!}{
    \begin{tabular}{lccc}
        \hline
        Method & Prompts/Sample & Tokens/Prompt & Tokens/Sample \\
        \hline
        CoT & 1 & 103.25 & 103.25 \\
        Self-Reflection & 3 & 255.86 & 767.59 \\
        Self-Consistency & 7 & 103.25 & 722.73 \\
        MAD & 10 & 276.72 & 2767.18 \\
        Self-Contrast & 6 & 349.80 & 2098.81 \\
        CFMAD (Ours) & 10 & 246.82 & 2468.22 \\
        \hline
    \end{tabular}
    }
    }
    \caption{Token Costs of different methods.}
    \label{tab:token_costs}
\end{table}

\subsection{Results on Open-Ended Tasks}

In addition to evaluating binary prediction and multi-choice QA tasks, where candidate answers are provided, we also conducted evaluations on more open-ended tasks, including mathematical reasoning and machine translation. Specifically, we selected three mathematical reasoning datasets—SVAMP~\citep{SVAMP}, GSM8K~\citep{GSM8K}, and Multi\_Arith~\citep{multiarith}—and one machine translation dataset, CommonMT~\citep{commonsenseMT}. These datasets do not offer predefined answer choices and require the language models (LLMs) to generate open-ended responses.

Since these tasks lack candidate answers, we first used CoT prompts to generate candidate answers from the open answer spaces by sampling multiple times. This allowed us to transform the problems into a multi-choice QA format, which could then be processed using our pipeline.

For evaluation, we used precision as a metric for the mathematical reasoning tasks and BLEURT1 for the machine translation task. The results, as shown in the Table~\ref{tab:open-ended}, demonstrate that CFMAD outperforms the other baselines, highlighting the effectiveness and generality of our approach.

\begin{table}[ht]
    \centering
    \setlength{\tabcolsep}{0.8mm}{
    \resizebox{\columnwidth}{!}{
    \begin{tabular}{lcccc}
        \hline
        Method & SVAMP & GSM8K & Multi\_Arith & CommonMT \\
        \hline
        CoT & 0.8067 & 0.7567 & 0.9575 & 0.7008 \\
        Self-Consistency & 0.8500 & 0.8167 & 0.9804 & \slash \\
        MAD & 0.8567 & 0.8100 & 0.9575 & 0.7183 \\
        Self-Contrast & 0.8467 & 0.8133	 & 0.9771 & 0.7119 \\
        CFMAD & \textbf{0.8700} & \textbf{0.8300}	& \textbf{0.9837} & \textbf{0.7201} \\
        \hline
    \end{tabular}
    }
    }
    \caption{Results on Open-Ended Tasks}
    \label{tab:open-ended}
\end{table}

\subsection{Results on Open-Sourced LLMs}

In addition to GPT-3.5, we have also assessed our method on two other open-source LLMs: Llama2-7B~\citep{llama2} and Llama3-8B~\citep{llama3}. The results on the CommonsenseQA dataset are presented in Table~\ref{tab:open-sourced}. Our approach, CFMAD, consistently outperforms other baselines across these models, highlighting the effectiveness of our method when applied to open-source LLMs.

\begin{table}[ht]
    \centering
    \setlength{\tabcolsep}{0.8mm}{
    \resizebox{\columnwidth}{!}{
    \begin{tabular}{lcc}
        \hline
        CommonsenseQA & Llama2-7B & Llama3-8B \\
        \hline
        Self-Consistency & 0.6233 & 0.7533 \\
        MAD & 0.5900 & 0.7467  \\
        Self-Contrast & 0.6233 & 0.7600 \\
        CFMAD & \textbf{0.6367} & \textbf{0.7700} \\
        \hline
    \end{tabular}
    }
    }
    \caption{Results on Open-Sourced Tasks}
    \label{tab:open-sourced}
\end{table}

\section{Experiments Details}

\subsection{Dataset Details} \label{datesetDetails}

We performed experiments using four datasets: Hover, BoolQ, CosmosQA, and CommonsenseQA. The details of these datasets are as follows:

\begin{itemize}[leftmargin=*]

    \item \textbf{Hover:} Hover is a fact-checking task dataset. Each instance in the Hover dataset consists of a claim and supporting evidence. The task requires multi-hop reasoning based on the supporting evidence to determine whether the evidence supports the claim or not.  

    \item \textbf{BoolQ:} BoolQ is a reading comprehension task dataset that consist of questions that can be answered with a simple ``yes'' or ``no''. And each question is paired with a paragraph from Wikipedia that contains the answer.

    \item \textbf{CosmosQA:} CosmosQA is a dataset focused on reading comprehension and commonsense reasoning. Each instance consists of a context and a question with four answer options that require inference beyond the text, using commonsense knowledge to determine the correct answer.

    \item \textbf{CommonsenseQA}: CommonsenseQA is a challenging dataset that tests a model's ability to use commonsense knowledge to answer multiple-choice questions. Each question has one correct answer and four distractors.

\end{itemize}

In this work, we first tested all 3-hop and 4-hop instances in the validation set of Hover, with 1,835 instances for 3-hop and 1,039 instances for 4-hop to demonstrate our method's effectiveness. Next, due to budget constraints, we randomly selected 300 instances from the validation set of each of the remaining three datasets to conduct our experiments. 

\subsection{Method Implementation Details}

For MAD, we initialized 3 agents and conducted 3 rounds of debate. For Self-Contrast, we had the LLM initially generate answers from 3 perspectives for subsequent contrast. For Self-Consistency, we initially generated 7 answers, voting for the final answer. For our CFMAD framework, we initially preset two predetermined answers to instruct the LLMs to generate abduction. For datasets like CosmosQA and CommonsenseQA, which have multiple potential answers, we first use 3 rounds of CoT prompting to obtain one potentially correct answer as a predetermined answer. Then, we randomly select another predetermined answer from the remaining options.

\section{Basline Prompts} \label{prompt1}

\subsection{CoT Prompt}

\begin{itemize}[leftmargin=*]

\item {
\textbf{Fact Check Task} \\
\textit{Evidence: \{evidence\} \\
Claim: \{claim\} \\
You are a fact checker. Please fully understand the evidence and claim, and answer is the claim true or false? Let us verify step by step. }
}

\item {
\textbf{Commonsense Resoning} \\
\textit{
\{Question and option here\} \\
Play the role of a common sense reasoning expert. Choose the most appropriate answer for the question. You are expected to explain your reasoning process step-by-step before providing the final answer.
Output format:\\
Reasoning steps: [Your precise reasoning steps here]\\
Judgement: The correct answer is Option [X].
}
}

\end{itemize}

\subsection{Relfection Prompt}

\begin{itemize}[leftmargin=*]

\item {
\textbf{Reflection Prompt} \\
\textit{As a critic, review the assistant's response. Identify any incorrect or missing information, and provide feedback. \\
\{Question Content Here\} \\
Assistant's reply: \{CoT\_reply\} \\
Output format: \\
Judgement: [Critically evaluate the assistant's response.] \\
Potential Improvements: [Suggest ways to enhance the accuracy or clarity of the assistant's response.]}
}

\item {
\textbf{Revision Prompt} \\
\textit{\{Question Content Here\} \\
Assistant's reply: \{CoT\_reply\} \\
Feeback: \{reflection\_reply\} \\
Based on the feedback provided, revise your response to the question. \\
Output format: \\
The correct answer is Option [X].}
}

\end{itemize}

\subsection{MAD}

Here we show the prompt for CommonsenseQA. The prompt structure is similar for other tasks, and the specific prompts for other tasks can be found in our code.

\begin{itemize}[leftmargin=*]

\item {
\textbf{Initial Prompt 1} \\
\textit{\{Question Content Here\} \\
Play the role of a common sense reasoning expert. Choose the most appropriate answer for the question. You are expected to explain your reasoning process step-by-step before providing the final answer.}
}

\item {
\textbf{Initial Prompt 2} \\
\textit{\{Question Content Here\} \\
Which option is the most appropriate answer based on the common sense?}
}

\item {
\textbf{Initial Prompt 3} \\
\textit{\{Question Content Here\} \\
Let us think step by step and find the most appropriate answer based on the common sense.}
}

\item {
\textbf{Debate Prompt} \\
\textit{\{Question Content Here\} \\
Let us think step by step and find the most appropriate answer based on the common sense. \\
Assistant: \{Your previous response\} \\
Other agent1: \{Other agents' previous responses1\} \\
Other agent2: \{Other agents' previous responses2\} \\
Using the judgements from other agents as additional information, can you give an updated response.}
}

\item {
\textbf{Judge Prompt} \\
\textit{\{Question Content Here\} \\
Let us think step by step and find the most appropriate answer based on the common sense. \\
Agent1: \{last response of agent 1\} \\
Agent2: \{last response of agent 2\} \\
Agent3: \{last response of agent 3\} \\
Three agents have given their answers. \\ According to the majority of the answers, what is the most appropriate answer? Your answer should look like this: ``The correct answer is Option [X]''}
}

\end{itemize}

\subsection{Self-contrast}
Here we show the prompt for CommonsenseQA. The prompt structure is similar for other tasks, and the specific prompts for other tasks can be found in our code.

\begin{itemize}[leftmargin=*]

\item {
\textbf{Self-Curate Prompt} \\
\textit{You are a commonsense reasoning specialist. You need to complete multiple choice questions related to commonsense reasoning. Given a question, you need to carefully analyze the question and dynamically generate several useful prompt instructions. These prompt instructions should be diverse and also useful for commonsense reasoning. These prompt instructions are used to guide the language model to think in different ways, attention to different emphases, and reason from different perspectives for more accurate commonsense reasoning. \\
For instance, you can adopt multi-faceted thinking (logical thinking, lateral thinking, analogical thinking, etc .), different reasoning perspectives( e.g., top-down, bottom-up , step-by-step), and different emphases of concern, (entity words, numbers, time, etc ) for input question in prompt instruction. \\
Here are some guidance rules for Prompt Generation: \\
1. Tone Requirement: Please generate prompt instructions in the third person. \\
2. Content Requirement: Each prompt instruction should adopt a different way of thinking, or focus on a different perspective, or different emphases to solve the question. \\
3. Number Requirement: Dynamically generate the most valuable 3 prompt instructions based on the input math question. \\
4. Format Requirement: Each prompt instruction should start with \#\#\# and end with @@@ \\
5. Others: Prompt instructions should focus on commonsense reasoning. So don't ask any other irrelevant questions in the prompt. \\
Here is an example : The question is: Who is the first president of the United States? \\
Output: \\
bottom - up perspective : \#\#\# As a specialist in commonsense reasoning, you have to judge the given question from a bottom-up perspective. Breaking the question down into smaller components or details. What specific pieces of information are provided in the question, and how do they contribute to understanding the problem? @@@ \\
The input question is: \{question\}. Please generate the most suitable three prompts:}
}

\item {
\textbf{Contrast Prompt} \\
\textit{You are a specialist in commonsense reasoning. Given some candidate judgements for a question, you should carefully compare the difference for each two judgements in their reasoning steps.\\
When you compare, you need to consider the following questions: \\
1: Are the two judgements have different final judge and judge reasons? \\
2: Where are the differences in their reason steps and judge reasons? \\
3. Why are the answers of the two judgements different? \\
After contrasting , you should generate a checklist based on these differences between candidate judgements . You should carefully consider each discrepancy and the reasons behind it, summarizing them into a few checking instructions in the checklist. This checklist can guide others to re-examine the input question and these candidate judgements to eliminate these discrepancies . \\
\{Question Content Here\} \\
Judgements:  \\
Judgement1: \{reply1\}, \\
Judgement2: \{reply2\}, \\
Judgement3: \{reply3\} \\
Output Format: \\
For Judgement1 and Judgement2 : [Give the difference between Judgement1 and Judgement2 here] \\
For Judgement1 and Judgement3 : [Give the difference between Judgement1 and Judgement3 here] \\
For Judgement2 and Judgement3 : [Give the difference between Judgement2 and Judgement3 here] \\
Checklist : [Give the directives for checking here]}
}

\item {
\textbf{Reflection Prompt} \\
\textit{Given a question, multiple inconsistent judgements, their differences in their reasoning processes and a checklist. You should revise the inconsistent reasoning step for each judgements, eliminate the differences, and output a new judgement. \\
Guidance Rules for Reflection: \\
1. Please check carefully according to the requirements on the checklist. It helps you to resolve conflicts between different judgements. \\
2. When you finish revising inconsistent judgements, please ensure all revised judgements should have the same answer . If not , please revise again until all inconsistencies are removed , and all candidates are consistent. \\
\{Question Content Here\} \\
The candidate judgements and their discrepancy are as follows: \\
\{ \\
    ``Candidate'': \{ \\
        ``Judgement'': ``\{reply1\}'', \\
        ``Judgement'': ``\{reply2\}'', \\
        ``Judgement3'': ``\{reply3\}'' \\
    \}, \\
    ``Discrepancy'': \{ \\
        ``difference\_1\_2'': \{ \\
            ``source'': ``Judgement1'', \\
            ``target'': ``Judgement2'', \\
            ``relation'': \{difference\_1\_2\} \\
        \}, \\
        ``difference\_1\_3'': \{ \\
            ``source'': ``Judgement1'', \\
            ``target'': ``Judgement3'', \\
            ``relation'': \{difference\_1\_3\} \\
        \}, \\
        ``difference\_2\_3'': \{ \\
            ``source'': ``Judgement2'', \\
            ``target'': ``Judgement3'', \\
            ``relation'': \{difference\_2\_3\} \\
        \} \} \} \\
Checklist: \{checklist\} \\
Please revise each inconsistent judgement and give your final judgement. \\
Output Format: \\
The answer is Option [X].}
}

\end{itemize}

\section{Our Prompts} \label{prompt2}

\subsection{Fact Check Task}

\begin{itemize}[leftmargin=*]
\item {
\textbf{Abduction Generation} \\
\textit{Evidence: \{evidence\} \\
Claim: \{claim\} \\
Please fully understand the evidence and claim, and answer why the claim is \{true/false\}?}
}

\item {
\textbf{Counterfactual Debate for Critic} \\
\textit{Evidence: \{evidence\} \\
Claim: \{claim\} \\
Assistant: \{reply of assistant\} \\
The Assistant's answer maybe wrong. Please persuade the assistant that the claim is actually incorrect based on the evidence.}
}

\item {
\textbf{Counterfactual Deabte for Assistant} \\
\textit{Evidence: \{evidence\} \\
Claim: \{claim\} \\
Please fully understand the evidence and claim, and answer why the claim is true? \\
Fact checker: \{reply of assistant\} \\
Critic: \{reply of crtic\} \\
Play the role of fact checker. Please point out the errors in critic's answer and reiterate your point.}
}

\item {
\textbf{Judge} \\
\textit{Evidence: \{evidence\} \\
Claim: \{claim\} \\
\{Debate Process for each stance\} \\
After hearing the positive and negative sides, do you think the claim is true or false? [True/False]}
}
\end{itemize}

\subsection{Commonsense Reasoning}

\begin{itemize}[leftmargin=*]
\item {
\textbf{Abduction Generation} \\
\textit{\{Question Content Here\} \\
Try to explain why the question's answer might be option \{predetermined answer\}. \\
Output Format: \\
Judgement: The answer is option \{predetermined answer\}. \\
Reasoning: [Your reasoning here]}
}

\item {
\textbf{Counterfactual Debate for Critic} \\
\textit{\{Question Content Here\} \\
Assistant: \{reply of assistant\} \\
The Assistant's answer maybe wrong. Please persuade the assistant that his answer maybe wrong.}
}

\item {
\textbf{Counterfactual Deabte for Assistant} \\
\textit{\{Question Content Here\} \\
Assistant: \{reply of assistant\} \\
Critic: \{reply of critic\} \\
As assistant, please refute the critic's answer and persuade the critic that your answer is correct.}
}

\item {
\textbf{Judge} \\
\textit{\{Question Content Here\} \\
Which option is the answer of the question? The results of the analysis for each of the possible options are as follows: \\
\{Debate Process for each stance\} \\
After seeing the debate process above, do you think which option is the most appropriate answer for the question? Please only give a correct answer and no other replies. \\
Output format: \\
Judgement: The correct answer is Option [X]. \\
Reasoning steps: [Your precise reasoning steps here]}
}
\end{itemize}

\end{document}